\documentclass[11pt,a4paper]{article}
\usepackage{times,latexsym}
\usepackage{url}
\usepackage[T1]{fontenc}
\usepackage{booktabs}
\usepackage{comment}
\usepackage{graphicx} 
\usepackage{hyperref}
\usepackage[acceptedWithA]{tacl2021v1}
\usepackage{tacl2021v1}
\usepackage{xspace,mfirstuc,tabulary}

\newif\iftaclinstructions
\taclinstructionsfalse % AUTHORS: do NOT set this to true
\iftaclinstructions
\renewcommand{\confidential}{}
\renewcommand{\anonsubtext}{(No author info supplied here, for consistency with
TACL-submission anonymization requirements)}
\newcommand{\instr}
\fi

\iftaclpubformat % this "if" is set by the choice of options

\else

\fi

\newcommand{\Nchildren}{14,006}
\newcommand{\Nitemsdbase}{872} 
\newcommand{\Nitemsremovedchildren}{10} 
\newcommand{\NitemsremovedLLMs}{312}
\newcommand{\NitemsremovedXLMV}{250}
\newcommand{\Nitems}{622}
\newcommand{\bertje}{BERTje}
\newcommand{\robbert}{RobBERT}
\newcommand{\xlmv}{XLM-V}
\newcommand{\gpttwo}{D-GPT-2}
\newcommand{\mgpt}{M-GPT}
\newcommand{\gptthree}{GPT-3}

\title{Do large language models solve verbal analogies like children do?}

\author{
   Claire E. Stevenson$^\diamond$,
   Mathilde ter Veen$^\diamond$,
   Rochelle Choenni$^\dagger$, \\
   \textbf{Han L. J. van der Maas}$^\diamond$
   and
   \textbf{Ekaterina Shutova}$^\dagger$
   \ \\
   $^\diamond$Psychological Methods, University of Amsterdam, the Netherlands
   \ \\
   $^\dagger$ILLC, University of Amsterdam, the Netherlands
   \ \\
   \texttt{c.e.stevenson@uva.nl}
 }

\date{}

\begin{document}
\maketitle
\begin{abstract}
Analogy-making lies at the heart of human cognition. Adults solve analogies such as \textit{Horse belongs to stable like chicken belongs to …?} by mapping relations (\textit{kept in}) and answering \textit{chicken coop}. In contrast, children often use association, e.g., answering \textit{egg}. This paper investigates whether large language models (LLMs) solve verbal analogies in A:B::C:? form using associations, similar to what children do. We use verbal analogies extracted from an online adaptive learning environment, where \Nchildren{} 7-12 year-olds from the Netherlands solved \Nitems{} analogies in Dutch. The six tested Dutch monolingual and multilingual LLMs performed around the same level as children, with \mgpt{} performing worst, around the 7-year-old level, and \xlmv{} and \gptthree{} the best, slightly above the 11-year-old level. However, when we control for associative processes this picture changes and each model's performance level drops 1-2 years. Further experiments demonstrate that associative processes often underlie correctly solved analogies. We conclude that the LLMs we tested indeed tend to solve verbal analogies by association with C like children do.
\end{abstract}

\section{Introduction}
Analogy-making, using what you know about one thing to infer knowledge about a new, somehow related instance, lies at the heart of human intelligence and creativity and forms the core of educational practice \citep{gentner1988, hofstadter1997, holyoak2012analogy}. Given how important analogical reasoning is to learning and generalization, much research has focused on how this seemingly unique human ability emerges, develops, and can be improved \citep{goswami1991, sternberg1980, stevenson2018} as well as emulated in machines \citep{gentner2011computational, mitchell2021}. Recently, large language models (LLMs), such as \gptthree{} \citep{brown2020language}, have demonstrated surprisingly good performance in verbal analogy solving (e.g., \textit{table is to legs as tree is to …? chair, leaves, branches or roots?}) \citep{Lu2022, webb2023emergent}. The question then arises how LLMs solve these analogies. Is the process similar to adult humans using relational mapping? Or perhaps more similar to the associative processes children tend to use? 

Earlier work shows that language models largely rely on semantic similarity between analogy terms to solve analogies \citep{rogers2020, ushio2021identifyanalogies}, which would indicate an associative process. In this paper we investigate whether six LLMs use similar associative processes to solve a set of Dutch children's verbal analogies. First, we examine how LLM performance compares to children and find that the best models, \gptthree{} and \xlmv{} \citep{liang2023xlm} perform around the same level as 11-year-olds. Second, we examine whether LLM performance is influenced by the same item characteristics that affect children's analogy solving, where results showed that \gptthree's performance follows a similar pattern to children's, whereas \xlmv's performance is more stable across item characteristics. Third, through a series of prompting experiments we show that these LLMs appear to use associative solving processes, similar to children, where performance drops 1-2 years in age level when we control for associative processes. 

This paper contributes to the study of analogical reasoning in LLMs in three ways: (1) it is the first to directly compare LLM analogy solving performance to that of children; (2) we use experiments to tap into the \textit{process} of analogy-making and compare this to human processes; and (3) we use Dutch rather than English language items and examine performance in mono- and multilingual LLMs.

\section{Theoretical Background}
\subsection{The Analogical Reasoning Process}
 Although there are different cognitive models of analogical reasoning -varying in the order of processing steps and whether these occur sequentially or in parallel, there is a general consensus on which processes are involved. Taking the example of “body is to feet as tree is to …?” (or more abstractly, A:B::C:?), the basic analogy information processing steps are generally considered to be: (1) encoding the relevant information about the base (A:B) and target (C) domains; (2) searching and retrieving relationships and similarities between the analogy elements in the base domain, A and B (e.g., “stands on” for body and feet); (3) aligning the base and target domains ("body and tree are things that stand") and mapping the mostly likely relationship between A and B, to the target domain, C, to come up with D; and (4) evaluating the validity of the predicted solution \citep{gentner2017analogy, sternberg1977, thibaut2016analogical}.

\subsection{Factors Affecting People's Verbal Analogy Solving}
This analogy solving process of encoding, finding relationships, alignment and mapping is consistently found in people from about 12 years and up \citep{thibaut2016analogical}. When adults make mistakes there are three main factors that to lead to errors: (1) more difficult types of relations to be mapped (e.g., causal versus categorical) (2) a large conceptual distance between analogy base and target domains, and (3) salient distractors amongst the multiple-choice options \citep{jones2022}. 

\paragraph{Type of Relation}
\citet{jones2022} grouped analogical relations into three types: categorical, causal and compositional. They found that adults perform better on categorical analogies (e.g., tarantula:spider::bee:insect) than causal (e.g., fracture:cast::incision:scar) or compositional (e.g., fingernail:finger::knee:leg) analogies. Children’s performance follows a similar pattern, assuming sufficient domain knowledge is in place \citep[e.g.,][]{sternberg1980, goswami1990melting, Alexander1991}.

\paragraph{Conceptual Distance Between Base and Target Domains}
The greater the distance between an analogy base and target domain the more difficult the analogy is for adults and children to solve \citep{jones2022, thibaut2016analogical}. For example, bowl:dish::spoon:silverware is easier for people to solve than wrench:tool::sad:mood. 

\paragraph{Distractor Salience}
People are sometimes lured to choose a distracting incorrect response in multiple choice verbal analogies, and are most easily distracted by answer options that have a strong semantic association with the C term \citep{kucwaj2022}. \citet{jones2022} defines distractor salience as the relation between C:D relative to each of the C:D’, where D’ represents each distractor option. Distractor salience is high, when the semantic similarity between C and one of the incorrect answers D’ is greater than the semantic similarity between C and the correct answer D. High distractor salience leads to lower performance in adults \citep{ichien2020, jones2022} and this is even more apparent in children \citep{Richland2006, thibaut2016analogical}.

\subsection{Analogical Reasoning Development} 
Children’s verbal analogical reasoning improves with age, where a gradual shift occurs around 4-8 years of age from reasoning based on surface similarities and associations to reasoning based on (abstract) relations \citep{gentner1988, stevenson2018, gentile1977}. For example, if we ask a four-year-old “Horse belongs to stable like chicken belongs to …?” they may use association and reply “egg”, relying on the strong connection between the words chicken and egg to solve the problem. In contrast, older children and adults will likely give the intended relational response “chicken coop”, using the underlying relation structure to solve the analogy.

Two main factors that seem to affect the transition from associative to relational reasoning are increased domain knowledge \citep{goswami1990melting, gentner1988, Alexander1991} and improved executive functions \citep[working memory and inhibition control;][]{doumas2018, thibaut2016analogical}. 

Children tend to fail in analogy solving if they are unfamiliar with the elements or relations in the analogy \citep{gentner2017analogy, goswami1990melting, goddu2020}. If children are shown to possess the required domain knowledge and are provided clear instructions on how to solve the task then they can successfully solve verbal analogies (in the form of pictures) as early as 3-years-old \citep{goswami1991, goddu2020}. 

However, even when children can solve these analogies, evidence from scene analogy problems \citep{Richland2006} and eye-tracking studies \citep{thibaut2016analogical} shows that children up to 8 years-old tend to focus first on the C term when solving analogies, sometimes ignoring A and B altogether \citep{thibaut2016analogical}. This appears to be related to limited working memory capacity \citep{Richland2006, stevenson2013wm, stevenson2017ai} and limits in inhibition- and executive control \citep{thibaut2016analogical, doumas2018}. Performance improves when interventions are used that support children's processing capacities \citep{stevenson2018} and when children are forced to focus first on the A:B pair \citep{glady2017}.

So what does the analogy solving process look like in children? According to the current literature, children up to about 8 years old, tend to initially rely on associative reasoning to solve analogies, specifically focusing on the C term and selecting its nearest associate. Because executive control processes improve with age, they are increasingly able to inhibit associative responses and take more abstract or complex relations into account when choosing their solution. When instructions are clear, required domain knowledge is present and the cognitive load is low, this can be as early as 3-years-old. However, associative responses remain the fallback strategy when relations are unfamiliar or processing capacities are overtaxed until about 8-years-old \citep{stevenson2018}.

\subsection{Verbal Analogy Solving in LLMs}
The extent to which LLMs can solve analogies is a subject of debate. Most of this work has focused on comparing models in terms of overall accuracy on benchmarks such as the Bigger Analogy Test Set \citep[BATS;][]{mikolov2013linguistic} and verbal analogies from the Scholastic Assessment Test \citep[SAT;][]{turney2003} and investigating the types of relations they can solve (e.g., syntactic versus semantic). More importantly, when LLMs demonstrate analogy solving abilities (e.g., \citet{webb2023emergent}, it is unclear which processes underlie the achieved solutions. In this study we take the unique perspective of relating LLM verbal analogy solving to human processing and development; to get a better understanding of the what and why of LLM verbal analogy solving.

\paragraph{Word embeddings}
A decade ago, \citet{mikolov2013linguistic} published their seminal paper showing that pre-trained word embeddings \citep[e.g., Word2Vec][]{mikolov2013distributed} could be used to solve verbal analogies in the form of A:B::C:? using vector arithmetic, the most famous example being: $embed(king) - embed(man) + embed(woman) \approx embed(queen)$, where $embed$ represents the word embedding obtained from the pre-trained neural network. This milestone was tempered by \citet{gladkova2016analogy}, who made clear that this method was limited in the breadth of relations that it could process. For example, the capitol-country relation was solved quite successfully, but others such as animal-sound and part-whole, were solved less successfully.

\paragraph{Transformer language models}
With the rise of the Transformer architecture, featuring language models such as BERT \citep{devlin2018}, verbal analogy solving by language models remained a challenge. Earlier work transferred the verbal analogy datasets, such as the BATS to the sentence level, and showed that BERT-based models and GPT-2 \citep{radford2019gpt2} performed at a similar level to GloVe \citep{pennington2014glove}, a word embedding model, on analogies containing relations such as capitol-country and male–female pairs \citep{zhu2020sentence}. More recently, \citet{czinczoll2022scientific} developed a dataset containing scientific and metaphor analogies. Here there was a clear advantage of transformer models over analogy solving with word embeddings, where GPT-2, BERT and M-BERT outperformed GloVe on the analogy items containing metaphors such as career:mountain::success:ascent. Yet, the general conclusion remained that verbal analogy solving is still a challenge for LLMs.

\paragraph{People versus LLMs in analogy solving}
Recent research has shown that LLMs can solve verbal analogies with similar accuracy to people. For example, \citet{ushio2021identifyanalogies} showed that LLMs such as GPT-2 and RoBERTa generally perform well on analogies designed for 4th to 10th graders (9-16 year-olds). Also, \citet{webb2023emergent} concluded that \gptthree{} generally performs around the same level as adults on four verbal analogy datasets. 

\paragraph{Item factors affecting LLM verbal analogy solving}
There has been some research on the effect of \textit{relationship type} on LLM's verbal analogy solving performance. \citet{ushio2021distillingrelations} showed that fine-tuned RoBERTa models performed slightly better on categorical relations (hypernymns) than compositional ones (meronymns). And \citet{webb2023emergent} found that categorical relations in the SAT verbal analogies were easier for \gptthree{} than compositional (function) relations and also that categorical relations were easier than both compositional and causal relations on the same items as those administered in \citet{jones2022}. Similarly, \citet{linford2022impact} found that categorical relations were  easier for BERT models than causal relations, although performance on both was far lower than for human adults.

Similarly to people, LLMs have more difficulty when the \textit{conceptual distance} between the domains in the analogy are far rather than near. For example, the LLMs in \citet{czinczoll2022scientific} performed better on the BATS analogies than on their SCAN dataset comprising scientific and metaphor based analogies, where the semantic distance between the base and target domains where greater for the SCAN dataset. In addition the scientific analogies were solved better by LLMs than those based on metaphors, which was explained by there being a clearer correspondence between base and target domains in scientific analogies. Also, \citet{webb2023emergent}, used the items from \citet{jones2022} to investigate whether, like in people, a near conceptual distance between the base and target domains made analogies easier to solve for \gptthree{} than far analogies; this was indeed the case. Interestingly, humans outperformed \gptthree{} on the far analogies. Similarly, \citealt{linford2022impact} found causal relations more difficult than categorical relations for the BERT and BART models they tested.

There is little research on the effect of \textit{distractor salience} on LLM analogy solving. However, \citet{ushio2021identifyanalogies} did conduct an experiment to evaluate to what degree the entire context of the analogy was needed for LLMs to solve analogies, by masking the head or tail of the candidate analogy pair. They found that RoBERTa and BERT only dropped 10 to 15 percentage points in accuracy, still achieving accuracies of 30\% or higher on the SAT analogies. On the one hand, this means that items may be constructed in a way that makes the correct solution the most likely choice. On the other hand, this could also indicate that associative processes, such as semantic similarity, may be a shortcut for LLM analogy solving (which we revisit in our experiments described below). Either way, we can expect that salient distractors, i.e.\ multiple-choice options that are semantically more similar to the analogy terms than the correct response, will have a greater chance of being "selected" by the LLMs.

\section{Research Questions}
In this pre-registered study (see \url{https://osf.io/g9c4j}) we examine how six LLMs' (Dutch monolingual and multilingual models: \robbert{} \citep{delobelle2020robbert}, \bertje{} \citep{liu2019roberta}, \xlmv{} \citep{liang2023xlm}, \gpttwo{} \citep{radford2019gpt2}, \mgpt{} \citep{shliazhko2022mgpt}, and \gptthree{} \citep{brown2020language}) solve verbal analogies extracted from a children's online adaptive learning environment.\footnote{We also tested GPT-4's ability to solve these analogies. GPT-4 was able to solve >70\% with only the C term (so without the A:B relation) and we concluded that data contamination from previously testing these same analogies a number of times in \gptthree{} and ChatGPT-3.5-turbo was the likely cause.}

\paragraph{RQ1: How well do LLMs perform compared to children ages 7-12 in verbal analogy solving?} Based on the literature, we expect recent LLMs to solve the analogies with similar accuracy to older children (12-year-olds) as this is similar to adult performance \citep[hypothesis 1a;][]{webb2023emergent, ushio2021distillingrelations}. We also expect that recent larger models, also trained on more data (e.g., \xlmv{} and \gptthree{} versus older BERT and GPT models) will perform better than older, smaller models trained on less data as this provides more domain knowledge to draw upon (hypothesis 1b)\footnote{The remaining RQs were investigated with the best performing GPT model and the best performing BERT model}. 

\paragraph{RQ2: Which item characteristics influence children's and LLM performance on verbal analogies?} We expect the pattern of results found in adults also to be found in children and in LLMs. First, we expect performance on categorical relations to be better than compositional and causal relations for both children \citep[][hypothesis 2a1]{sternberg1980} and LLMs \citep[][hypothesis 2a2]{webb2023emergent}. Second, we expect analogies with a near conceptual distance between A:B to be easier than far analogies for children (\citet{thibaut2016analogical}; Hypothesis 2b1) and LLMs \citep[][hypothesis 2b2]{czinczoll2022scientific, webb2023emergent}. Third, we expect higher distractor salience to lead to far more errors in children, because distractors that are semantically more similar to the C term than the correct answer are the most likely choice when solving an analogy based on association with C \citep[][hypothesis 2c1]{thibaut2016analogical}. We expect association to be the main mechanism by which LLMs solve verbal analogies and that they, similar to children, choose answer options with a strong semantic association to C \citep[][hypothesis 2c2]{ushio2021identifyanalogies}. 

\paragraph{RQ3: Do LLMs solve verbal analogies using associative processes or analogical reasoning?} We investigate this through a series of experiments comparing LLM performance on alternative formations of the verbal analogies, where we control for associative reasoning.

\section{Methods}
LLM data and code are publicly available on \url{https://osf.io/zsh2v/}. Children's data is available upon request.

\subsection{Prowise Learn's Verbal Analogies Game}
Prowise Learn is an online adaptive learning environment for elementary school children. Over 2,000 elementary schools and >200,000 children in the Netherlands use Prowise Learn have practiced their reading, writing and arithmetic skills both in- and outside of school. 

The verbal analogies game is one game among a series of reading and writing games in the "Language Sea" game space. Figure 1 contains a screenshot of the verbal analogies game, where an analogy in the form of "A:B::C:?" is presented in written form and the children must choose among five answer options, all five of which are semantically associated with C.

\begin{figure}
\caption{Language Sea example analogy "lawyer : defending :: teacher : educating"}
\centering
\includegraphics[width=0.5\textwidth]{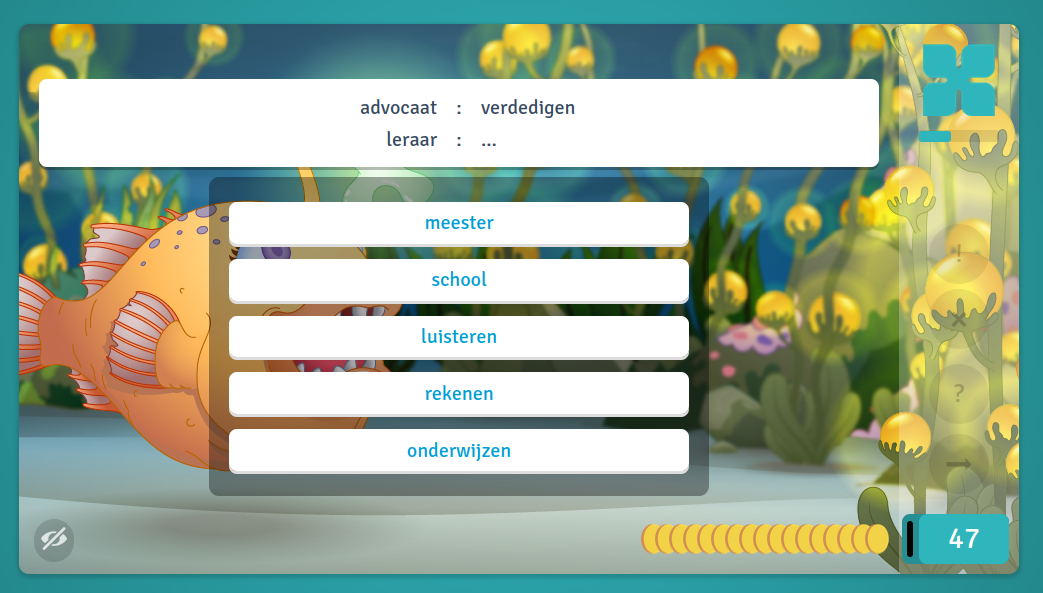}
\end{figure}

Prowise Learn games are adaptive, so that children solve items that are neither too difficult nor too easy, presenting children with items that they have a 65-85\% chance of solving correctly, using response time to improve estimations of their ability \citep{klinkenberg2011}. Each time a child solves an item his/her ability score on the game is updated according to an algorithm similar to the adaptive ELO rating system used for chess players \citep[for details see][]{klinkenberg2011}. At the same time the item's difficulty level is adapted according to the same algorithm. In this way item difficulty is on the same scale as the children's ability, and, as such item difficulties can be used to study children's abilities \citep[see][for examples in the math, language and logical reasoning domains]{vdMaas2015multiplication, vdMaas2017voice, vdMaas2013mastermind}. 

The ELO adaptive item presentation algorithm is based on the one-parameter logistic function from item response theory where we estimate the probability a child will solve an item correctly given the child's ability score $\theta$ and the item's difficulty level $\beta$ as shown in Equation 1. 
\begin{equation}
P(X=1|\theta,\beta)=\frac{e^{(\theta-\beta)}}{1+e^{(\theta-\beta)}}
\end{equation}
If a child's ability score is higher than the item's difficulty then they will have a higher probability of solving the item correctly; when $\theta$ and $\beta$ are equal then the probability of solving the item correctly is .5.

\paragraph{Data Collection with Children}
For this study, we extracted information on \Nchildren{} 7-12 year-old's (M = 10.73, SD = 1.15 years) performance on \Nitemsdbase{} verbal analogies from the Prowise Learn database. We applied three selection criteria when extracting the children's data (on June 19, 2021): (1) children solved at least 20 items in order to ensure we had stable ability estimates, (2) children had last played the game on or after September 1st 2020, the start of the school year and 4 months after the launch of the game, when item difficulty estimates were verified to have small standard errors\footnote{The item difficulties range from -11.27 to 8.14, the mean item difficulty estimate is -1.91 and the mean SE of this estimate is 4.63.} and (3) children were ages 7-12 to avoid confounds in performance (i.e., younger children most likely did not have sufficient reading abilities and older children had most likely repeated a grade). 

\paragraph{Item Selection}
The game contained three types of verbal reasoning problems; verbal analogies was one of them. From the initial set of \Nitemsdbase{} verbal analogies, we removed items that: (1) contained words/tokens in the analogy or in two or more response options that were missing from the \xlmv{} vocabulary (removed \NitemsremovedXLMV) \footnote{We decided to deviate from our pre-registration to only remove items that \xlmv{} processed in a way that was incompatible with our probing method (\NitemsremovedXLMV) rather that all items that at least one LLM could not process (\NitemsremovedLLMs) so as to have the largest possible item set to conduct our analyses on.} and (2) were judged by two independent raters to contain errors (e.g., multiple correct solutions, requiring domain knowledge that children are most likely unfamiliar with; \Nitemsremovedchildren{} items removed\footnote{We only checked items with >1.5 SD item difficulty ratings for errors.}). This resulted in \Nitems{} for data analysis.

\paragraph{Items for Experiment 4}
In our last experiment to investigate whether the LLMs indeed chose associated distractors above the correct response we adapted a selection of 34 items from the original item set. First, they were transposed to the form A:C::B:D (e.g., key:keyboard::page:book transposed to key:page::keyboard:book). Second, since the original distractors were related to the C term, we selected new distractors related to the B term using Small World of Words \cite{smallworldwords}, a project where native speakers list words that immediately come to mind when prompted with a stimulus word. The first and second author examined the listed associations and selected four distractor options that made sense given the context of the analogy, but could not be considered correct solutions (e.g., in the original form "paper, reading cover and title" were the distractors, in the transposed version "computer, typing, pc and piano" were selected as distractors).

\paragraph{Information extracted per item}
The following information was extracted per item: question text, answer options, item difficulty rating, standard error of the item difficulty rating, type of analogy relation, number of times the item was solved, proportion of times each response option was selected. 

\subsection{Item characteristics}
\paragraph{Relation Type}
Relationship type is defined as how the A and B term are related. This relationship must be applied to the C term to find D. Table \ref{tab:analogyexamples} provides an overview of the types of relations in the Prowise Learn analogies game\footnote{These labels were chosen and annotated by the Prowise Learn item developers, Education or Psychology graduates with experience in item development.}. For analyses related to RQ2 we selected 302 items that fall into the following three categories defined by \citet{jones2022}: 
\begin{itemize}
  \item \textbf{Categorical}: one of the A:B terms defines the category and the other word is an example of this category. For example “yellow” is part of the category “color”. 
  \item \textbf{Causal}: one of the A:B terms is the cause and the other is the effect. For example “stumbling” will result in “falling”.  
  \item \textbf{Compositional}: one of the A:B terms is part of the other term. For example: “leaf” is part of a “tree”.
\end{itemize} 

\begin{table*}[ht]
\centering
\begin{tabular}{lrll}
  \hline
Prowise Learn relations & N &  relations$^{*}$ & example \\ 
  \hline
 action-result &  36 & causal & parasol : shadow :: sun : warmth \\ 
   classification &  51 & categorical & lego : toys :: sock : clothes \\ 
   part-whole &  51 & compositional & gate : city :: door : house \\ 
   problem-solution &   6 & causal & noisy : earplugs :: illness : medicine \\ 
   share characteristic &  25 & compositional & giant : mountain :: dwarf : mouse \\ 
   same category &  28 & categorical & lion : tiger :: dog : wolf \\ 
   lacks aspect &  18 &  & naked : clothing :: bald : hair \\ 
   belong together &  38 &  & evening : dinner :: midday : lunch \\ 
   item-characteristic &  45 & compositional & skyscraper : high :: lead : heavy \\ 
   degree$|$strength &  24 &  & warm : hot :: strange :: crazy \\ 
   object-function &  34 & compositional & pan : cooking :: pen : writing \\ 
   object-location &  46 &  & bracelet : arm :: necklace : neck \\ 
   cause-effect &  11 & causal & falling : broken :: heating : hot \\ 
   synonyms &  41 &  & crying : whining :: sending : delivering \\ 
   opposites &  51 &  & remembering: forgetting :: heating : cooling \\ 
   indicates &  27 &  & mushrooms : autumn :: cold : winter \\ 
   actor-action &  29 &  & dog : wagging :: cat : purring \\ 
   \hline
\end{tabular}
\caption{$^{*}$ Mapping of relations in verbal analogies game to those examined in \citet{jones2022}.}
\label{tab:analogyexamples}
\end{table*}

\paragraph{Conceptual Distance Between Base and Target Domains}\label{conceptualdistance}
We used three vector-based language models\footnote{Word2Vec trained by CLIPS on different Dutch corpora \citep{tulkens2016}, Word2Vec trained by the Nordic Language Processing Laboratory on the CoNLL17 corpus \citep{kutuzov2017}, and FastText trained on Common Crawl and Wikipedia \citep{grave2018}.} to compute the semantic distance (cosine distance) between the A:B and the C:D pair. We categorized the distances of these three vector models as near (distance ranging from 0-.35), middle (.36-.64) or far distance (.65-1.0). Then we determined the semantic distance between base and target (near or far) per item and per model and used the most frequent category (near or far) as the category for analysis. Items in the middle category were dropped for this analysis (217 items removed).

\paragraph{Distractor Salience}
Distractor salience was measured by the cosine similarity between C and D minus the cosine similarity between C and each incorrect answer D'. The distractor salience is high when the similarity between C and D' is higher than the similarity between C and the correct answer \cite{jones2022}. We used the same three vector-based models mentioned in Section~\ref{conceptualdistance} to get the vector representations of the words and compute the cosine distances between C and each of the five D's. Then we determined distractor salience (high or low) per item for each vector model and used the most frequent category (high or low) as the category for analysis.

\subsection{Prompting with LLMs}
\paragraph{Pretrained Language Models}
We studied how six different pretrained LLMs solved the same set of verbal analogies as the children. Three of the LLMs are monolingual and 3 multilingual, all of which rely on the Transformer architecture \citep{vaswani2017attention}. 

Three of our models are BERT-based masked language models: \textbf{\bertje{}} \citep{de2019bertje} and \textbf{\robbert{}} \citep{delobelle2020robbert} that are both pretrained on Dutch data only, and \robbert's  multilingual variant \textbf{\xlmv{}} \citep{liang2023xlm} that was trained on 116 languages.\footnote{Given that we only include single-token words, we found \xlmv{} to be more suitable than mBERT or XLM-R as it suffers less from overtokenization in Dutch and thus covers more of our test words.}  Similar to their English counterparts, \robbert{} is the robustly optimized version of BERTje\footnote{Note that BERTje replaces Next Sentence Prediction with the Sentence Order Prediction objective from \citet{lanalbert}} that uses SentencePiece instead of WordPiece for tokenization and only uses Masked Language Modelling as a pretraining task. Moreover, identical to BERT \citep{kenton2019bert}, all three models contain 12 layers with 12 attention heads each.

The other three models are autoregressive Transformer-decoder based language models. The Dutch version of GPT-2 which we refer to as \textbf{\gpttwo{}} \citep{de2021good} exploits the pretrained GPT-2 model \citep{radford2019gpt2} and only retrains the lexical embeddings in order to account for a Dutch vocabulary. Whereas, \textbf{\mgpt{}} \citep{shliazhko2022mgpt} is the multilingual version of GPT-2 that was pretrained from scratch on 60 different languages. We also use \textbf{\gptthree{}} \citep{brown2020language}, specifically the 'text-davinci-003' engine.

\paragraph{Elicitation and Scoring Methods}
We wanted to mimic the way the children solved the analogy items in the best way possible. This was especially important because we are investigating whether an associative response is more likely in the presence of a correct response. Therefore, we prompted \gptthree{} with the full analogy and asked it to choose from the five response options. For example, "tripping is to falling as picking up is to ? Choose clean, junk, mess, room, or thrift store." The response options were presented in random order. 

This method was not possible for the BERT-based models and led to unexpected output for earlier GPT models. Therefore, we used the masked language model approach and fed the models `\texttt{A} is to \texttt{B}, as \texttt{C} is to \texttt{D}', replacing \texttt{D} with each possible multiple-choice solution. The \texttt{D} option with the highest probability for the completion was considered the selected response.

To account for variation in query quality, we designed 5 slightly different prompt templates in Dutch, all similar to `\texttt{A} is to \texttt{B}, as \texttt{C} is to \texttt{D}', to retrieve analogy completions. This means that we administered each item five times (once for each template). We report the mode of the scores using multiple templates, reducing the effect of context and taking into account that the best template differed for the models (see Section~\ref{templateacc}).

Please note that due to tokenization issues some analogy terms or response options could not be processed by the models using our prompting method. We report accuracies for items where all of the analogy terms and at least two distractor response options could be processed. We also report the number of items the accuracy score was based on.

\label{templateacc}
\begin{table*}
    \footnotesize
    \centering
    \begin{tabular}{l|cccccc}
    \toprule
    Template &	RoBERT & Bertje & \xlmv & \gpttwo & \mgpt & \gptthree \\
    \midrule
    N items & 697 & 672 & 639 & 382 & 115 & 872 \\
    \midrule
    A staat tot B, zoals C staat tot D & 0.53 & 0.47 & 0.55 & \textbf{0.43} & \textbf{0.36} & \textbf{0.55} \\
    A hoort bij B, zoals C hoort bij D & 0.53 & 0.46 & 0.56 & 0.42 & 0.33 & 0.54 \\
    A is vergelijkbaar met B, zoals C vergelijkbaar is met D & 0.49 & 0.46 & 0.51 & 0.40 & \textbf{0.36} & 0.53 \\
    A is tot B, zoals C is tot D & \textbf{0.55} & \textbf{0.48} & \textbf{0.59} & 0.38 & 0.32 & \textbf{0.55} \\
    A hoort bij B op dezelfde manier dat C hoort bij D & 0.51 & 0.44 & 0.56 & 0.40 & 0.31 & 0.50 \\
    \bottomrule
    \end{tabular}
    \caption{Overview of accuracy score for each template for each LLM. Highest accuracy is marked in \textbf{bold}.}
\end{table*}

\paragraph{Word embeddings}
We also included two word embedding models, i.e. Word2Vec \citep{tulkens2016} and FastText \citep{grave2018}, to establish a baseline by completing the analogy with vector arithmetic using the formula C - A + B  where the selected D is the highest ranked outcome of the five answer options \citep{mikolov2013linguistic}.

\section{Results RQ1: How well do LLMs perform compared to children?}

\begin{figure}
    \centering
    \includegraphics[width=\linewidth]{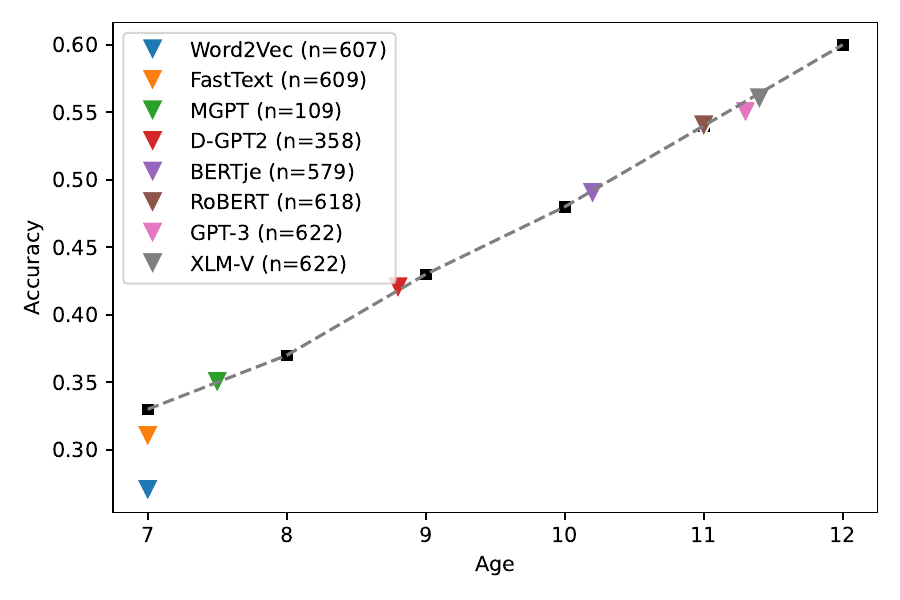}
    \caption{At what age level does each LLM perform?}
    \label{fig:modelages}
\end{figure}

Figure \ref{fig:modelages} shows performance per model on the \Nitems{} items. Here we see that see that 7-year-olds outperform the baseline performance of vector models, but that the six LLMs we tested outperform 7-year-olds. We also see that the best performing models (\robbert, \xlmv{} and \gptthree) have around the same level of accuracy as 11-year-olds, confirming hypotheses 1a and 1b. 

Given these results we proceed with answering our RQs using the best performing GPT-based model, \gptthree, and the best BERT-based model, \xlmv. 

\section{Results RQ2: Which item factors influence analogy solving?}
For RQ2, we tested the effects of solver (children, \gptthree, and \xlmv) and/or item characteristics on accuracy using logistic regression.

\begin{figure}
    \centering
    \includegraphics[width=\linewidth, height=5.5cm]{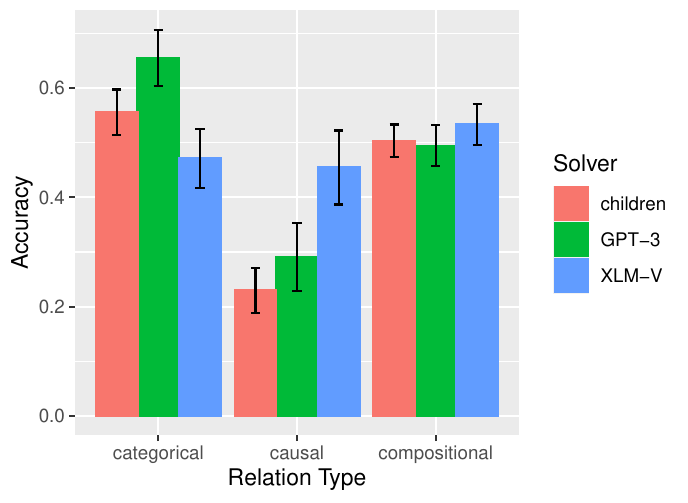}
    \caption{\gptthree{} follows the adult pattern, where categorical relations are easiest, followed by compositional relations and causal are most difficult. \xlmv{} solves items of all three relations equally well.}
    \label{fig:relations}
\end{figure}

\paragraph{Relation Type}
We expected to find categorical relations to be the easiest followed by compositional and then causal relations for both children (H2a) and LLMs (H2b).
As can be seen in Figure \ref{fig:relations}, \gptthree's results follow this pattern, where categorical relations were solved better than compositional relations ($z=2.45, p=.014$) and causal were solved worse ($z=-2.61, p=.009$). In children, causal relations are more difficult that compositional relations ($z=-3.46, p<.001$), but there are no reliable differences in performance on categorical versus compositional relations ($z=0.79, p=.429$). In contrast, \xlmv{} performed similarly on all three relation types (categorical versus compositional: $z=0.79, p=.429$, causal versus compositional: $z=0.79, p=.429$). Interestingly, \xlmv{} performed better than children and \gptthree{} on causal relations. This is surprising as BERT-based LLMs seem to have difficulty with causal relations \citep[e.g.,][]{bhagavatula2019abductive}. However, it is no surprise that children, like adults, find these relations the most difficult \citealp{jones2022}.

\begin{figure}
    \centering
    \includegraphics[width=\linewidth, height=5.5cm]{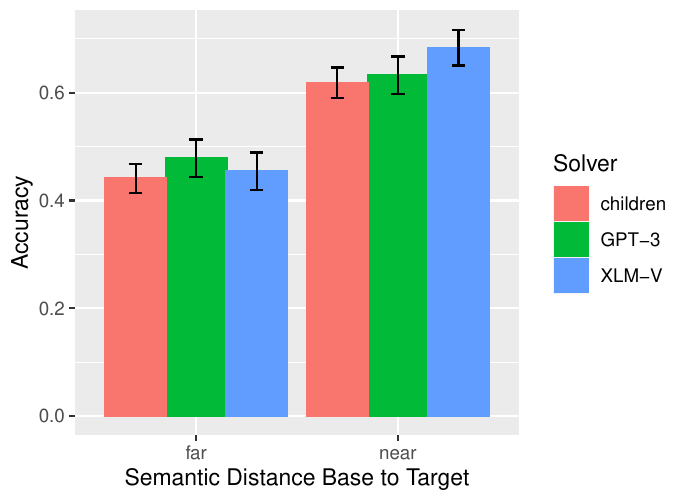}
    \caption{Near analogies are easier to solve than far analogies.}
    \label{fig:nearfar}
\end{figure}

\paragraph{Near vs Far Distance between Base and Target Domains}
As can be seen in Figure~\ref{fig:nearfar}, items with a near semantic distance between the base and target domains were easier for both children and LLMs than those with a far semantic distance, confirming hypothesis H2b ($z=3.55, p<.001$).

\begin{figure}
    \centering
    \includegraphics[width=\linewidth, height=5.5cm]{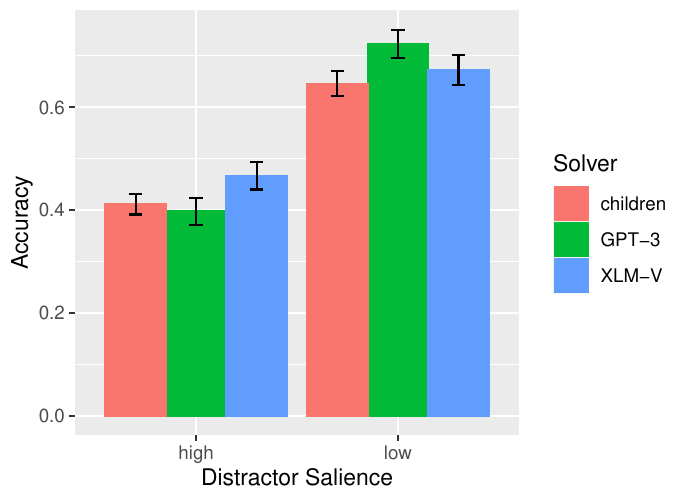}
    \caption{Analogies with low distractor salience are easier to solve than those with high distractor salience.}
    \label{fig:highlow}
\end{figure}

\paragraph{Distractor Salience}
Items with lower distractor salience were easier to solve than those with high distractor salience for both children and LLMs (see Figure~\ref{fig:highlow}; $z=5.66, p<.001$), confirming H2c. 

\section{Results RQ3: Do LLMs use associative or analogical processes to solve analogies?}
We conducted a series of experiments using alternative formulations of the analogies to investigate how LLMs solve the verbal analogies, explicitly testing and controlling for associative responses such as those seen in younger children.

\subsection{Experiment 1: C:?} In experiment 1, we prompt the LLMs with only the C term, e.g., "C is to [MASK]". If these are solved by association as we expect, then LLMs should still be able to solve a substantial portion of analogies purely by association with C \citep{ushio2021identifyanalogies, poliak2018hypothesis}; hypothesis 3a). This was indeed the case, where \gptthree{} solved 33\% of the items with only the C term and \xlmv{} solved 28\% with only the C term. 

For both \gptthree{} and \xlmv, the items that were solved with C:? as a prompt were in all three relation categories and a near or far conceptual distance to the A:B pair was irrelevant. However, distractor salience had a clear effect, where those with low distractor salience were more often solved correctly with only C:? than those with highly salient distractors (\gptthree: $z=5.52, p<.001$, \xlmv: $z=6.84, p<.001$).

\begin{figure}
    \centering
    \includegraphics[width=\linewidth, height=5.5cm]{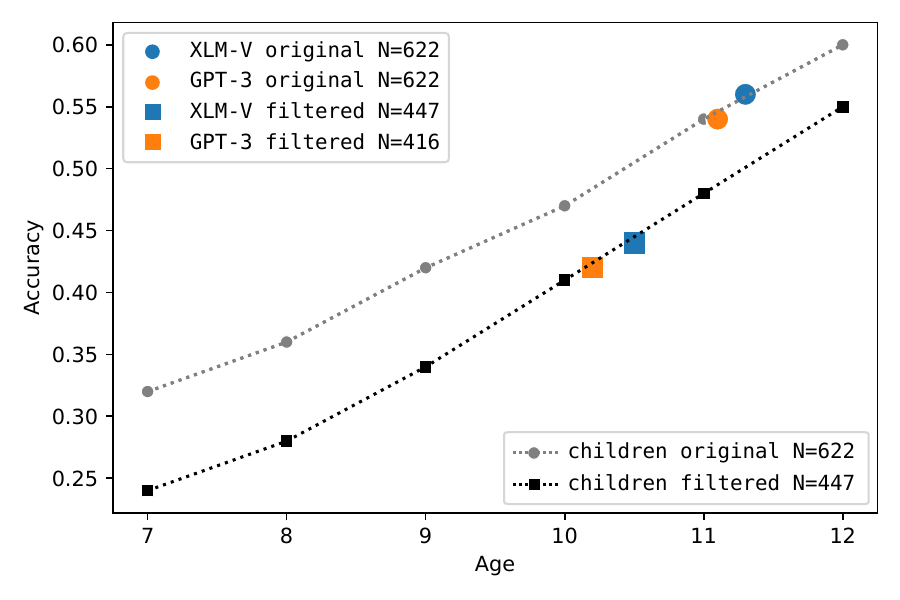}
    \caption{Experiment 2. Model performance drops a year if we filter out items solved correctly with C:?. Children's accuracy also decreased, but to a lesser degree (11-year-olds -.06 accuracy versus -.12 for both \xlmv{} and \gptthree).}
    \label{fig:dat_noC}
\end{figure}

\subsection{Experiment 2: A:B::C:? for selected items} 
We then removed the items that each model solved with only the C term and reevaluated their performance compared to children on the remaining items (see Figure \ref{fig:dat_noC}). In both cases, the models performance dropped to below that of the average child, and to around the 10 year-old-level. 

\begin{table}[ht]
\centering
\begin{tabular}{lcccc}
  \hline
   Exp & Problem & N items & \xlmv & \gptthree \\
   \midrule
   0 & A:B::C:? & \Nitems & 0.56 & 0.55\\
   1 & C:? & \Nitems & 0.28 & 0.33 \\
   2 & A:B::C:? & 237 & 0.44 & 0.42 \\
   3 & A:C::B:? & \Nitems & 0.49 & 0.50 \\
   4 & A:C::B:? & 34 & 0.29 & 0.38 \\
   \hline
\end{tabular}
    \caption{LLM performance on the four experiments.}
\label{tbl:experiments}
\end{table}

\subsection{Experiment 3: A:C::B:?} 
Here we present the same verbal analogy items, but transposed to the form of "A is to C as B is to ?". By switching the B and C terms this precludes the analogy from being solved purely by association as all of the distractor options are related to the original C term and not B. We expected this to result in better LLM performance than in experiment 2 because the full analogy is available and there are no distractors that complete the analogy by association with B (\citet{ushio2021identifyanalogies}; hypothesis 3b). 

\gptthree{} performed at nearly the same level on the transposed analogies as the original analogies (original .55, transposed .50) and \xlmv's performance was only slightly lower (original .56, transposed .49). This was expected (H3b) given that the analogies involved the same words, only in a different order, and the distractors were designed to be related to the C-term, not the B-term.

In both cases, performance was somewhat better than in experiment 2, where associative responses were ruled out (see Table~\ref{tbl:experiments}). It is possible that the LLMs were forced to extract the relation between A and C, but it is also possible that LLM performance solely improved because distractors were not related to the B-term. To control for this we conducted one last experiment.

\subsection{Experiment 4: A:C::B:? for selected items} Here we again present verbal analogy items in the form of "A is to C as B is to ?". But, now we selected items that were solved correctly in both A:B::C:? and A:C::B:? forms, but incorrectly in C:? form. This led to a selection of analogies that were robustly solved correctly by both LLMs, but not solved purely by association (as C:?). We then changed the distractor options to be associated to B. And removed items that could be solved as B:? without the A:C relation, leaving us with 34 items.
We stipulated that if performance again degraded then there is further evidence that LLMs rely on association rather than relational mapping to solve these verbal analogies. 

As can be seen in Table \ref{tbl:experiments}, our results showed that performance dropped substantially, albeit still above chance level (20\%).

\subsection{RQ4: Do the LLMs make the same errors children do?}
We also conducted an exploratory analysis to see if LLMs made the same mistakes children did, i.e., choosing the same erroneous distractor option. We selected items that were answered incorrectly by \gptthree{} and \xlmv{} and then compared their most likely chosen distractor to the distractor that most children chose. \gptthree{} chose the same distractor as most children 39\% of the time, whereas for \xlmv{} the most probable incorrect response was the same as most children in 33\% of cases. For example, in the analogy "pencil is to pencil case, as sweater is to ?", children and LLMs most often chose "pants" (wrong answer) instead of "closet" (right answer). But, on other items such as "stem is to plant as snout is to ?", children and \gptthree{} chose "paws" (wrong answer), whereas XLM-V's most likely response was "mouse" (right answer). There were not many cases in which both LLMs chose the same erroneous distractor, highlighting the differences in the two language models.

\section{Discussion}
The main goal of this paper was to investigate whether LLMs tend to use association to solve verbal analogies, similar to what children do, instead of relational mapping like adults. Direct comparison of performance between the children and LLMs showed that the best performing LLMs, \xlmv{} and \gptthree{}, scored at around the 11 year-old level, an age when we expect children to have transitioned to adult-like analogical reasoning \citep{thibaut2016analogical}. However, the more we controlled for associative responses, the greater the models' performance degraded. Our experiments provide evidence that association rather than relational reasoning likely underlies the process by which LLMs solve verbal analogies. LLMs are, in essence, next (or missing) word prediction machines \citep{mccoy2023embers}. Thus, it is no surprise that our research shows that these LLMs tend to respond with the most associated next word (i.e., the model's statistically most likely next word). This finding falls in line with the more general research that questions whether reasoning actually occurs or that LLMs are more like Searle's "Chinese Rooms" that pass on the right output because this is what they were trained to do \citep{mccoy2023embers, zevcevic2023causalparrots, wu2023reciting}. 

However, previous research concluded that analogical reasoning has likely emerged in \gptthree{} \citep{webb2023emergent}. The question then arises why LLM performance on this children's verbal analogy task is not as strong as their performance on other verbal analogy tasks and benchmarks where they seem to outperform even adult college students \citep[e.g.,][]{webb2023emergent}. We see three main explanations. First, the items were in Dutch and not English which means that the models may have less training on the analogy terms. We tried to account for this by removing items where \xlmv{} could not process part of the analogy given the way we probed the model. But, explicit testing of domain knowledge both in children and LLMs would be advisable for future research. Second, our items contained four distractor options rather than the traditional one distractor used by \citet{webb2023emergent} and \citet{jones2022}. This reduces the chance of choosing the correct option from 50\% to 20\%, making our test more robust \citep{mitchell2023smartAI}. Third and finally, our items were designed to lure children into associative reasoning. All of the distractors were associated with the C-term. The other tests were designed for adults and may not have provided distractors that LLMs would be "lured to choose".

Although our findings point towards LLMs choosing associative solutions similar to what children choose to solve the verbal analogies, does this mean that they solve the analogies like children do? We examined how relation type, conceptual distance between the analogy base and target and the salience of the distractors influenced the LLMs and children's performance. \gptthree's performance was affected in the same way as children's on all accounts. As previous research with people and \gptthree{} shows, causal relations were more difficult than compositional ones and categorical items were easiest \citep{jones2022, webb2023emergent}. And both \xlmv{} and \gptthree{}, like children, performed better on near analogies than far ones and solved items with low distractor salience better than those with highly salient distractors. When the LLMs made errors, their most likely response was the same as what most children were lured by about $1/3$ of the time. In sum, there is clearly some overlap between children and LLMs in which solutions were chosen, but further research still needs to uncover which mechanisms underlie why this is the case. 
\subsection{Conclusion}
The most important take-away point from our study is that LLMs may solve analogies as well as 11 year-olds, but to ascertain whether analogical reasoning is emerging in these systems we must take the \textit{mechanisms} by which they obtain these solutions into account. In contrast to \citet{webb2023emergent}, our findings provide evidence against emerging analogical reasoning based on relational mapping and point towards associative processes, perhaps similar to those in children.
\bibliography{verbalanalogies}

\begin{thebibliography}{62}
\expandafter\ifx\csname natexlab\endcsname\relax\def\natexlab#1{#1}\fi

\bibitem[{Alexander and Kulikowisch(1991)}]{Alexander1991}
P.~Alexander and J.~Kulikowisch. 1991.
\newblock \href {https://doi.org/10.1080/10862969109547735} {Domain knowledge and analogical reasoning ability as predicators of expository text}.
\newblock \emph{Journal of Reading Behavior}, 23(2):165--190.

\bibitem[{Bhagavatula et~al.(2019)Bhagavatula, Bras, Malaviya, Sakaguchi, Holtzman, Rashkin, Downey, Yih, and Choi}]{bhagavatula2019abductive}
Chandra Bhagavatula, Ronan~Le Bras, Chaitanya Malaviya, Keisuke Sakaguchi, Ari Holtzman, Hannah Rashkin, Doug Downey, Scott Wen-tau Yih, and Yejin Choi. 2019.
\newblock Abductive commonsense reasoning.
\newblock In \emph{International Conference on Learning Representations}.

\bibitem[{de~Bree et~al.(2017)de~Bree, van~der Ven, and van~der Maas}]{vdMaas2017voice}
Elise de~Bree, Sanne van~der Ven, and Han van~der Maas. 2017.
\newblock \href {https://doi.org/10.1080/15475441.2016.1217777} {The voice of holland: Allograph production in written dutch past tense inflection}.
\newblock \emph{Language Learning and Development}, 13(3):215--240.

\bibitem[{Brown et~al.(2020)Brown, Mann, Ryder, Subbiah, Kaplan, Dhariwal, Neelakantan, Shyam, Sastry, Askell et~al.}]{brown2020language}
Tom Brown, Benjamin Mann, Nick Ryder, Melanie Subbiah, Jared~D Kaplan, Prafulla Dhariwal, Arvind Neelakantan, Pranav Shyam, Girish Sastry, Amanda Askell, et~al. 2020.
\newblock Language models are few-shot learners.
\newblock \emph{Advances in neural information processing systems}, 33:1877--1901.

\bibitem[{Czinczoll et~al.(2022)Czinczoll, Yannakoudakis, Mishra, and Shutova}]{czinczoll2022scientific}
Tamara Czinczoll, Helen Yannakoudakis, Pushkar Mishra, and Ekaterina Shutova. 2022.
\newblock Scientific and creative analogies in pretrained language models.
\newblock \emph{arXiv preprint arXiv:2211.15268}.

\bibitem[{De~Deyne and Storms(2008)}]{smallworldwords}
S.~De~Deyne and G.~Storms. 2008.
\newblock Word associations: Norms for 1,424 dutch words in a continuous task.
\newblock \emph{Behavior Research Methods}, 40:198--205.

\bibitem[{De~Vries et~al.(2019)De~Vries, van Cranenburgh, Bisazza, Caselli, van Noord, and Nissim}]{de2019bertje}
Wietse De~Vries, Andreas van Cranenburgh, Arianna Bisazza, Tommaso Caselli, Gertjan van Noord, and Malvina Nissim. 2019.
\newblock Bertje: A dutch bert model.
\newblock \emph{arXiv preprint arXiv:1912.09582}.

\bibitem[{Delobelle et~al.(2020)Delobelle, Winters, and Berendt}]{delobelle2020robbert}
Pieter Delobelle, Thomas Winters, and Bettina Berendt. 2020.
\newblock Robbert: a dutch roberta-based language model.
\newblock In \emph{Findings of the Association for Computational Linguistics: EMNLP 2020}, pages 3255--3265.

\bibitem[{Devlin et~al.(2018)Devlin, Chang, Lee, and Toutanova}]{devlin2018}
Jacob Devlin, Ming-Wei Chang, Kenton Lee, and Kristina Toutanova. 2018.
\newblock Bert: Pre-training of deep bidirectional transformers for language understanding.
\newblock \emph{arXiv preprint arXiv:1810.04805}.

\bibitem[{Doumas et~al.(2018)Doumas, Morrison, and Richland}]{doumas2018}
Leonidas A.~A. Doumas, Robert~G. Morrison, and Lindsey~E. Richland. 2018.
\newblock \href {https://doi.org/10.3389/fpsyg.2018.01235} {Individual differences in relational learning and analogical reasoning: A computational model of longitudinal change}.
\newblock \emph{Frontiers in Psychology}, 9:1235.

\bibitem[{Gentile et~al.(1977)Gentile, Tedesco-Stratton, Davis, Lund, and Agunanne}]{gentile1977}
J.~R. Gentile, L.~Tedesco-Stratton, E.~Davis, N.~J. Lund, and B.~C. Agunanne. 1977.
\newblock \href {https://doi.org/10.1016/0160-2896(77)90019-8} {Associative responding versus analogical reasoning by children}.
\newblock \emph{Intelligence}, 1(4):369--380.

\bibitem[{Gentner(1988)}]{gentner1988}
D.~Gentner. 1988.
\newblock \href {https://doi.org/10.1111/j.1467-8624.1988.tb03194.x} {Metaphor as structure mapping: The relational shift}.
\newblock \emph{Child Development}, 59(1):47--59.

\bibitem[{Gentner and Forbus(2011)}]{gentner2011computational}
Dedre Gentner and Kenneth~D Forbus. 2011.
\newblock \href {https://doi.org/10.1002/wcs.105} {Computational models of analogy}.
\newblock \emph{Cognitive science}, 2(3):266--276.

\bibitem[{Gentner and Hoyos(2017)}]{gentner2017analogy}
Dedre Gentner and Christian Hoyos. 2017.
\newblock \href {https://doi.org/10.1111/tops.12278} {Analogy and abstraction}.
\newblock \emph{Topics in cognitive science}, 9(3):672--693.

\bibitem[{Gierasimczuk et~al.(2013)Gierasimczuk, van~der Maas, and Raijmakers}]{vdMaas2013mastermind}
Nina Gierasimczuk, Han~LJ van~der Maas, and Maartje~EJ Raijmakers. 2013.
\newblock An analytic tableaux model for deductive mastermind empirically tested with a massively used online learning system.
\newblock \emph{Journal of Logic, Language and Information}, 22(3):297--314.

\bibitem[{Gladkova et~al.(2016)Gladkova, Drozd, and Matsuoka}]{gladkova2016analogy}
Anna Gladkova, Aleksandr Drozd, and Satoshi Matsuoka. 2016.
\newblock \href {https://aclanthology.org/N16-2002.pdf} {Analogy-based detection of morphological and semantic relations with word embeddings: what works and what doesn’t.}
\newblock In \emph{Proceedings of the NAACL Student Research Workshop}, pages 8--15.

\bibitem[{Glady et~al.(2017)Glady, French, and Thibaut}]{glady2017}
Y.~Glady, R.~M. French, and J.-P. Thibaut. 2017.
\newblock \href {https://doi.org/10.3389/fpsyg.2017.00707} {Children’s failure in analogical reasoning tasks: A problem of focus of attention and information integration?}
\newblock \emph{Frontiers in Psychology}, 8:707.

\bibitem[{Goddu et~al.(2020)Goddu, Lombrozo, and Gopnik}]{goddu2020}
M.~K. Goddu, T.~Lombrozo, and A.~Gopnik. 2020.
\newblock \href {https://doi.org/10.1111/cdev.13412} {Transformations and transfer: Preschool children understand abstract relations and reason analogically in a causal task}.
\newblock \emph{Child Development}, 91(6):1898--1915.

\bibitem[{Goswami(1991)}]{goswami1991}
U.~Goswami. 1991.
\newblock \href {https://doi.org/10.2307/1130701} {Analogical reasoning: What develops? a review of research and theory}.
\newblock \emph{Child Development}, 62(1):1--22.

\bibitem[{Goswami and Brown(1990)}]{goswami1990melting}
Usha Goswami and Ann~L Brown. 1990.
\newblock Melting chocolate and melting snowmen: Analogical reasoning and causal relations.
\newblock \emph{Cognition}, 35(1):69--95.

\bibitem[{Grave et~al.(2018)Grave, Bojanowski, Gupta, Joulin, and Mikolov}]{grave2018}
E.~Grave, P.~Bojanowski, P.~Gupta, A.~Joulin, and T.~Mikolov. 2018.
\newblock \href {http://arxiv.org/abs/arXiv:1802.06893} {Learning word vectors for 157 languages}.

\bibitem[{Hofstadter(1997)}]{hofstadter1997}
Douglas~R Hofstadter. 1997.
\newblock \emph{Fluid concepts and creative analogies: computer models of the fundamental mechanisms of thought}.
\newblock Allen Lane, The Penguin Press.

\bibitem[{Holyoak(2012)}]{holyoak2012analogy}
Keith~J Holyoak. 2012.
\newblock Analogy and relational reasoning.
\newblock \emph{The Oxford handbook of thinking and reasoning}, pages 234--259.

\bibitem[{Ichien et~al.(2020)Ichien, Lu, and Holyoak}]{ichien2020}
Nicholas Ichien, Hongjing Lu, and Keith~J Holyoak. 2020.
\newblock \href {https://doi.org/10.3758/s13428-019-01312-3} {Verbal analogy problem sets: An inventory of testing materials}.
\newblock \emph{Behavior Research Methods}, 52(5):1803--1816.

\bibitem[{Jones et~al.(2022)Jones, Kmiecik, Irwin, and Morrison}]{jones2022}
Laura~L Jones, Matt~J Kmiecik, John~L Irwin, and Robert~G Morrison. 2022.
\newblock \href {https://doi.org/10.3758/s13423-022-02062-8} {Differential effects of semantic distance, distractor salience, and relations in verbal analogy}.
\newblock \emph{Psychonomic Bulletin \& Review}.

\bibitem[{Kenton and Toutanova(2019)}]{kenton2019bert}
Jacob Devlin Ming-Wei~Chang Kenton and Lee~Kristina Toutanova. 2019.
\newblock Bert: Pre-training of deep bidirectional transformers for language understanding.
\newblock In \emph{Proceedings of NAACL-HLT}, pages 4171--4186.

\bibitem[{Klinkenberg et~al.(2011)Klinkenberg, Straatemeier, and van~der Maas}]{klinkenberg2011}
Sylvia Klinkenberg, Marthe Straatemeier, and Han~L van~der Maas. 2011.
\newblock \href {https://doi.org/10.1016/j.compedu.2011.02.003} {Computer adaptive practice of maths ability using a new item response model for on the fly ability and difficulty estimation}.
\newblock \emph{Computers \& Education}, 57(2):1813--1824.

\bibitem[{Kucwaj et~al.(2022)Kucwaj, Ociepka, and Chuderski}]{kucwaj2022}
Hubert Kucwaj, Michał Ociepka, and Adam Chuderski. 2022.
\newblock \href {https://doi.org/10.3758/s13421-022-01285-3} {Various sources of distraction during analogical reasoning}.
\newblock \emph{Memory \& Cognition}, 50(7):1614--1628.

\bibitem[{Kutuzov et~al.(2017)Kutuzov, Fares, Oepen, and Velldal}]{kutuzov2017}
Andrey Kutuzov, Murhaf Fares, Stephan Oepen, and Erik Velldal. 2017.
\newblock Word vectors, reuse, and replicability: Towards a community repository of large-text resources.
\newblock \emph{Link{\"o}ping University Electronic Press}.

\bibitem[{Lan et~al.()Lan, Chen, Goodman, Gimpel, Sharma, and Soricut}]{lanalbert}
Zhenzhong Lan, Mingda Chen, Sebastian Goodman, Kevin Gimpel, Piyush Sharma, and Radu Soricut.
\newblock Albert: A lite bert for self-supervised learning of language representations.
\newblock In \emph{International Conference on Learning Representations}.

\bibitem[{Liang et~al.(2023)Liang, Gonen, Mao, Hou, Goyal, Ghazvininejad, Zettlemoyer, and Khabsa}]{liang2023xlm}
Davis Liang, Hila Gonen, Yuning Mao, Rui Hou, Naman Goyal, Marjan Ghazvininejad, Luke Zettlemoyer, and Madian Khabsa. 2023.
\newblock Xlm-v: Overcoming the vocabulary bottleneck in multilingual masked language models.
\newblock \emph{arXiv preprint arXiv:2301.10472}.

\bibitem[{Linford et~al.(2022)Linford, Ichien, Holyoak, and Lu}]{linford2022impact}
Bryce Linford, Nicholas Ichien, Keith Holyoak, and Hongjing Lu. 2022.
\newblock Impact of semantic representations on analogical mapping with transitive relations.
\newblock In \emph{Proceedings of the Annual Meeting of the Cognitive Science Society}, volume~44.

\bibitem[{Liu et~al.(2019)Liu, Ott, Goyal, Du, Joshi, Chen, Levy, Lewis, Zettlemoyer, and Stoyanov}]{liu2019roberta}
Yinhan Liu, Myle Ott, Naman Goyal, Jingfei Du, Mandar Joshi, Danqi Chen, Omer Levy, Mike Lewis, Luke Zettlemoyer, and Veselin Stoyanov. 2019.
\newblock Roberta: A robustly optimized bert pretraining approach.
\newblock \emph{arXiv preprint arXiv:1907.11692}.

\bibitem[{Lu et~al.(2022)Lu, Ichien, and Holyoak}]{Lu2022}
Hongjing Lu, Nicolas Ichien, and Keith~J Holyoak. 2022.
\newblock \href {https://doi.org/10.1037/rev0000358} {Probabilistic analogical mapping with semantic relation networks}.
\newblock \emph{Psychological Review}.

\bibitem[{McCoy et~al.(2023)McCoy, Yao, Friedman, Hardy, and Griffiths}]{mccoy2023embers}
R~Thomas McCoy, Shunyu Yao, Dan Friedman, Matthew Hardy, and Thomas~L Griffiths. 2023.
\newblock Embers of autoregression: Understanding large language models through the problem they are trained to solve.
\newblock \emph{arXiv preprint arXiv:2309.13638}.

\bibitem[{Mikolov et~al.(2013{\natexlab{a}})Mikolov, Sutskever, Chen, Corrado, and Dean}]{mikolov2013distributed}
Tomas Mikolov, Ilya Sutskever, Kai Chen, Greg~S Corrado, and Jeff Dean. 2013{\natexlab{a}}.
\newblock Distributed representations of words and phrases and their compositionality.
\newblock \emph{Advances in neural information processing systems}, 26.

\bibitem[{Mikolov et~al.(2013{\natexlab{b}})Mikolov, Yih, and Zweig}]{mikolov2013linguistic}
Tom{\'a}{\v{s}} Mikolov, Wen-tau Yih, and Geoffrey Zweig. 2013{\natexlab{b}}.
\newblock Linguistic regularities in continuous space word representations.
\newblock In \emph{Proceedings of the 2013 conference of the north american chapter of the association for computational linguistics: Human language technologies}, pages 746--751.

\bibitem[{Mitchell(2021)}]{mitchell2021}
Melanie Mitchell. 2021.
\newblock \href {https://doi.org/10.1111/nyas.14619} {Abstraction and analogy‐making in artificial intelligence}.
\newblock \emph{Annals of the New York Academy of Sciences}, 1505(1):79--101.

\bibitem[{Mitchell(2023)}]{mitchell2023smartAI}
Melanie Mitchell. 2023.
\newblock How do we know how smart ai systems are?

\bibitem[{Pennington et~al.(2014)Pennington, Socher, and Manning}]{pennington2014glove}
Jeffrey Pennington, Richard Socher, and Christopher~D Manning. 2014.
\newblock \href {https://nlp.stanford.edu/pubs/glove.pdf} {Glove: Global vectors for word representation}.
\newblock In \emph{Proceedings of the 2014 conference on empirical methods in natural language processing (EMNLP)}, pages 1532--1543.

\bibitem[{Poliak et~al.(2018)Poliak, Naradowsky, Haldar, Rudinger, and Van~Durme}]{poliak2018hypothesis}
Adam Poliak, Jason Naradowsky, Aparajita Haldar, Rachel Rudinger, and Benjamin Van~Durme. 2018.
\newblock \href {https://arxiv.org/pdf/1805.01042.pdf} {Hypothesis only baselines in natural language inference}.
\newblock In \emph{Proceedings of the 7th Joint Conference on Lexical and Computational Semantics}.

\bibitem[{Radford et~al.(2019)Radford, Wu, Child, Luan, Amodei, Sutskever et~al.}]{radford2019gpt2}
Alec Radford, Jeffrey Wu, Rewon Child, David Luan, Dario Amodei, Ilya Sutskever, et~al. 2019.
\newblock Language models are unsupervised multitask learners.
\newblock \emph{OpenAI blog}, 1(8):9.

\bibitem[{Richland et~al.(2006)Richland, Morrison, and Holyoak}]{Richland2006}
Lindsey~E. Richland, Robert~G. Morrison, and Keith~J. Holyoak. 2006.
\newblock \href {https://doi.org/10.1016/j.jecp.2006.02.002} {Children's development of analogical reasoning: Insights from scene analogy problems}.
\newblock \emph{Journal of Experimental Child Psychology}, 94(3):249--273.

\bibitem[{Rogers et~al.(2020)Rogers, Kovaleva, Downey, and Rumshisky}]{rogers2020}
Anna Rogers, Olga Kovaleva, Doug Downey, and Anna Rumshisky. 2020.
\newblock \href {https://doi.org/10.1609/aaai.v34i05.6398} {Getting closer to ai complete question answering: A set of prerequisite real tasks}.
\newblock In \emph{Proceedings of the AAAI Conference on Artificial Intelligence}, volume~34, pages 8722--8731.

\bibitem[{Shliazhko et~al.(2022)Shliazhko, Fenogenova, Tikhonova, Mikhailov, Kozlova, and Shavrina}]{shliazhko2022mgpt}
Oleh Shliazhko, Alena Fenogenova, Maria Tikhonova, Vladislav Mikhailov, Anastasia Kozlova, and Tatiana Shavrina. 2022.
\newblock mgpt: Few-shot learners go multilingual.
\newblock \emph{arXiv preprint arXiv:2204.07580}.

\bibitem[{Sternberg(1977)}]{sternberg1977}
Robert~J Sternberg. 1977.
\newblock \href {https://doi.org/10.1037/0033-295X.84.4.353} {Component processes in analogical reasoning}.
\newblock \emph{Psychological Review}, 84(4):353--378.

\bibitem[{Sternberg and Nigro(1980)}]{sternberg1980}
Robert~J Sternberg and Georgia Nigro. 1980.
\newblock Developmental patterns in the solution of verbal analogies.
\newblock \emph{Child Development}, 51:27--38.

\bibitem[{Stevenson(2017)}]{stevenson2017ai}
Claire~E Stevenson. 2017.
\newblock \href {https://link.springer.com/article/10.1007/s40593-016-0129-5} {Role of working memory and strategy-use in feedback effects on children’s progression in analogy solving: An explanatory item response theory account}.
\newblock \emph{International Journal of Artificial Intelligence in Education}, 27:393--418.

\bibitem[{Stevenson et~al.(2013)Stevenson, Heiser, and Resing}]{stevenson2013wm}
Claire~E Stevenson, Willem~J Heiser, and Wilma~CM Resing. 2013.
\newblock \href {https://doi.org/10.1016/j.cedpsych.2013.02.001} {Working memory as a moderator of training and transfer of analogical reasoning in children}.
\newblock \emph{Contemporary Educational Psychology}, 38(3):159--169.

\bibitem[{Stevenson and Hickendorff(2018)}]{stevenson2018}
Claire~E Stevenson and Marian Hickendorff. 2018.
\newblock \href {https://doi.org/10.1016/j.lindif.2018.04.010} {Learning to solve figural matrix analogies: The paths children take}.
\newblock \emph{Learning and Individual Differences}, 66:16--28.

\bibitem[{Thibaut and French(2016)}]{thibaut2016analogical}
Jean-Pierre Thibaut and Robert~M French. 2016.
\newblock Analogical reasoning, control and executive functions: a developmental investigation with eye-tracking.
\newblock \emph{Cognitive Development}, 38:10--26.

\bibitem[{Tulkens et~al.(2016)Tulkens, Emmery, and Daelemans}]{tulkens2016}
Stijn Tulkens, Chris Emmery, and Walter Daelemans. 2016.
\newblock Evaluating unsupervised dutch word embeddings as a linguistic resource.
\newblock Retrieved from http://hdl.handle.net/1854/LU-8507081.

\bibitem[{Turney et~al.(2003)Turney, Littman, Bigham, and Shnayder}]{turney2003}
Peter~D Turney, Michael~L Littman, Jeffrey Bigham, and Victor Shnayder. 2003.
\newblock \href {https://arxiv.org/pdf/cs/0309035.pdf} {Combining independent modules to solve multiple-choice synonym and analogy problems}.
\newblock In \emph{Proceedings of the International Conference on Recent Advances in Natural Language Processing (RANLP-03)}.

\bibitem[{Ushio et~al.(2021{\natexlab{a}})Ushio, Camacho-Collados, and Schockaert}]{ushio2021distillingrelations}
Akihiro Ushio, Jose Camacho-Collados, and Steven Schockaert. 2021{\natexlab{a}}.
\newblock \href {https://arxiv.org/pdf/2105.04949.pdf} {Distilling relation embeddings from pre-trained language models}.
\newblock In \emph{Proceedings of the 2021 Conference on Empirical Methods in Natural Language Processing}, pages 9044--9062. Association for Computational Linguistics.

\bibitem[{Ushio et~al.(2021{\natexlab{b}})Ushio, Espinosa-Anke, Schockaert, and Camacho-Collados}]{ushio2021identifyanalogies}
Akihiro Ushio, Luis Espinosa-Anke, Steven Schockaert, and Jose Camacho-Collados. 2021{\natexlab{b}}.
\newblock \href {https://doi.org/10.18653/v1/2021.acl-long.280} {Bert is to nlp what alexnet is to cv: Can pre-trained language models identify analogies?}
\newblock In \emph{Proceedings of the 59th Annual Meeting of the Association for Computational Linguistics and the 11th International Joint Conference on Natural Language Processing}, pages 3609--3624. Association for Computational Linguistics.

\bibitem[{Vaswani et~al.(2017)Vaswani, Shazeer, Parmar, Uszkoreit, Jones, Gomez, Kaiser, and Polosukhin}]{vaswani2017attention}
Ashish Vaswani, Noam Shazeer, Niki Parmar, Jakob Uszkoreit, Llion Jones, Aidan~N Gomez, {\L}ukasz Kaiser, and Illia Polosukhin. 2017.
\newblock Attention is all you need.
\newblock \emph{Advances in neural information processing systems}, 30.

\bibitem[{van~der Ven et~al.(2015)van~der Ven, Straatemeier, Jansen, Klinkenberg, and van~der Maas}]{vdMaas2015multiplication}
Sanne~HG van~der Ven, Marthe Straatemeier, Brenda~RJ Jansen, Sharon Klinkenberg, and Han~LJ van~der Maas. 2015.
\newblock \href {https://doi.org/10.1016/j.lindif.2015.08.013} {Learning multiplication: An integrated analysis of the multiplication ability of primary school children and the difficulty of single digit and multidigit multiplication problems}.
\newblock \emph{Learning and Individual Differences}, 43:48--62.

\bibitem[{de~Vries and Nissim(2021)}]{de2021good}
Wietse de~Vries and Malvina Nissim. 2021.
\newblock As good as new. how to successfully recycle english gpt-2 to make models for other languages.
\newblock In \emph{Findings of the Association for Computational Linguistics: ACL-IJCNLP 2021}, pages 836--846.

\bibitem[{Webb et~al.(2023)Webb, Holyoak, and Lu}]{webb2023emergent}
Taylor Webb, Keith~J Holyoak, and Hongjing Lu. 2023.
\newblock \href {https://www.nature.com/articles/s41562-023-01659-w} {Emergent analogical reasoning in large language models}.
\newblock \emph{Nature Human Behaviour}, 7:1526–1541.

\bibitem[{Wu et~al.(2023)Wu, Qiu, Ross, Aky{\"u}rek, Chen, Wang, Kim, Andreas, and Kim}]{wu2023reciting}
Zhaofeng Wu, Linlu Qiu, Alexis Ross, Ekin Aky{\"u}rek, Boyuan Chen, Bailin Wang, Najoung Kim, Jacob Andreas, and Yoon Kim. 2023.
\newblock Reasoning or reciting? exploring the capabilities and limitations of language models through counterfactual tasks.
\newblock \emph{arXiv preprint arXiv:2307.02477}.

\bibitem[{Ze{\v{c}}evi{\'c} et~al.(2023)Ze{\v{c}}evi{\'c}, Willig, Dhami, and Kersting}]{zevcevic2023causalparrots}
Matej Ze{\v{c}}evi{\'c}, Moritz Willig, Devendra~Singh Dhami, and Kristian Kersting. 2023.
\newblock Causal parrots: Large language models may talk causality but are not causal.
\newblock \emph{arXiv preprint arXiv:2308.13067}.

\bibitem[{Zhu and de~Melo(2020)}]{zhu2020sentence}
Xunjie Zhu and Gerard de~Melo. 2020.
\newblock \href {https://doi.org/10.18653/v1/2020.coling-main.300} {Sentence analogies: Linguistic regularities in sentence embeddings}.
\newblock In \emph{Proceedings of the 28th International Conference on Computational Linguistics}, pages 3389--3400.

\end{thebibliography}
\bibliographystyle{acl_natbib}

\onecolumn

\end{document}